# Hardware-Free Event Cameras Temporal Synchronization Based on Event Density Alignment

Wenxuan Li[1][0009-0006-3645-1374], Yan Dong[1][0000-0002-1428-6367], Shaoqiang Qiu[1][0009-0002-8681-2378] and Bin Han[1,2(✉)]

[1] State Key Laboratory of Intelligent Manufacturing Equipment and Technology, Huazhong University of Science and Technology, Wuhan 430074, Hubei, China
`binhan@hust.edu.cn`
[2] Guangdong Intelligent Robotics Institute, Dongguan, 523106, China

**Abstract.** Event cameras are a novel type of sensor designed for capturing the dynamic changes of a scene. Due to factors such as trigger and transmission delays, a time offset exists in the data collected by multiple event cameras, leading to inaccurate information fusion. Thus, the collected data needs to be synchronized to overcome any potential time offset issue. Hardware synchronization methods require additional circuits, while certain models of event cameras (e.g., CeleX5) do not support hardware synchronization. Therefore, this paper proposes a hardware-free event camera synchronization method. This method determines differences between start times by minimizing the dissimilarity of the event density distributions of different event cameras and synchronizes the data by adjusting timestamps. The experiments demonstrate that the method's synchronization error is less than 10ms under various senses with multiple models of event cameras.

## 1 Introduction

Event cameras are a type of neuromorphic sensor. Unlike traditional cameras that capture images at a fixed frame rate, event cameras can capture real-time changes in illumination. As a dynamic vision sensor, each pixel of an event camera can independently detect logarithmic relative lighting changes. Upon the lighting change of a pixel exceeding the set threshold, the event camera outputs event data [1-3]. Event cameras have several advantages, including no frame rate or minimum output time interval limitation, extremely low latency, and microsecond-level time resolution. They are suitable

This work is supported in part by the Guangdong Innovative and Entrepreneurial Research Team Program(2019ZT08Z780), in part by Dongguan Introduction Program of Leading Innovative and Entrepreneurial Talents (20181220), in part by the Natural Science Foundation of Hubei Province of China (2022CFB239), in part by the State Key Laboratory of Intelligent Manufacturing Equipment and Technology.

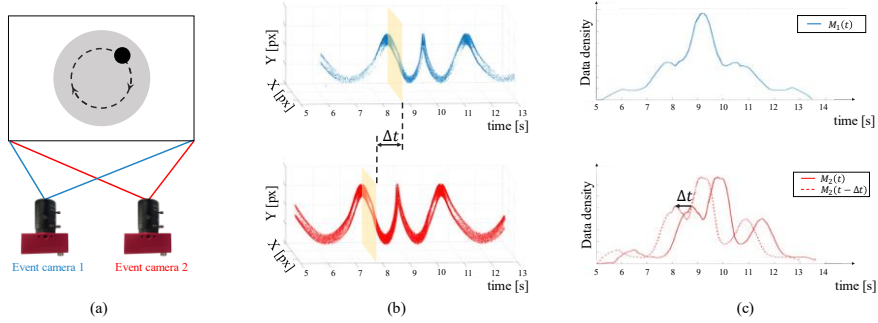

**Fig. 1.** An illustration of event stream alignment. (a) Two event cameras capture a rotating circle at varying angular velocities simultaneously. (b) The original event stream data of the two cameras, where $\Delta t$ is the difference between the start times of the two cameras. (c) $\Delta t$ is calculated by aligning the distributions of event densities $M(t)$.

for processing high-speed-moving scenes, e.g., racing cars and drones. Multiple event cameras can be applied to expand the visual field or acquire multi-view perspectives for a more comprehensive understanding of the environment.

Synchronization is necessary to ensure the accuracy and reliability of data in a multiple event camera system. As a result of trigger and transmission delays, start time differences between cameras lead to the time offset issues. Hardware synchronization is a commonly used approach to solve the problem. For instance, the DAVIS 240/346 camera created by iniVation[1] requires a stable reference voltage of 3.3V or 5V to generate special event trigger signals for synchronization, while Prophesee's products[2] rely on a 5V reference voltage. The increasing number of sensors results in a corresponding increase in hardware complexity which, in turn, negatively impacts the portability of the method. In addition, certain models of event cameras, such as CeleX5 from CelexPixel[3], do not support external synchronization. Therefore, a hardware-free synchronization method can achieve synchronization of multiple cameras of different models without the need for external circuits.

This paper proposes a novel hardware-free method to achieve synchronization among multiple event cameras by aligning their event density distributions. The method calculates the start time differences by minimizing the dissimilarity between the event density distributions of different cameras (see Fig. 1) and adjusts the timestamps accordingly. The proposed method achieves high accuracy in both indoor and outdoor environments, and can address the synchronization issue of various event cameras of different models.

---

[1] https://inivation.com/
[2] https://www.prophesee.ai/
[3] https://www.omnivision-group.com/technology/

## 2  Related Work

### 2.1  Hardware Synchronization

Certain models of event cameras support hardware synchronization. The hardware synchronization method commonly employs an external circuit platform to solve the time deviation problem. Calabrese et al. [4] synchronized the internal timestamps of the cameras by using a 3.5mm cable with a 10kHz clock to connect the cameras. Zou et al. [5] employed a specialized synchronization circuit to initiate the data acquisition of two event cameras and an RGB camera at the same time and prevent any time-shifting. Gao et al. [6] produced a trigger signal by using the onboard external oscillator as the primary clock and the STM32F407 microprocessor.

Hardware-based synchronization is typically performed manually or semi-automatically. However, in cases where the dataset involves long periods and multiple unsynchronized sensors, manual synchronization may prove challenging or impractical [8].

### 2.2  Hardware-Free Synchronization

Rueckauer et al. [7] utilized the DAVIS camera's integrated IMU to acquire more precise timestamps. The synchronization is accomplished via the motion pose detected by the IMU, yielding synchronization accuracy down to the microsecond level. Osadcuks et al. [8] proposed a clock-based synchronization method for data acquisition sensor array architecture to obtain microsecond-level synchronization with minimal hardware additions to the PIC32 microcontroller. Hu et al. [9] and Binas et al. [10] used computer time to approximately calibrate different sensors, and the errors generated thereby can be ignored due to the data transfer rate of only 10Hz. Zhu et al. [11] maximized the correlation between the DAVIS camera's integrated IMU and the VI sensor gyroscope angular velocity to calculate time offset. Post-processing can also achieve alignment if there is no synchronization during data acquisition. For instance, Censi et al. [12] suggested matching changes in image intensity with DVS event rates to identify temporal deviations between various data streams. However, these approaches are only applicable to synchronizing event cameras with other sensors and cannot accomplish synchronization among multiple event cameras.

Our method avoids the hardware complexity and compatibility issues associated with hardware synchronization, as well as the issue of some event cameras not supporting it. By calculating the differences between the start times of multiple event cameras through software, our method effectively addresses time deviation issues among multiple event cameras.

## 3 Methodology

### 3.1 Problem Description

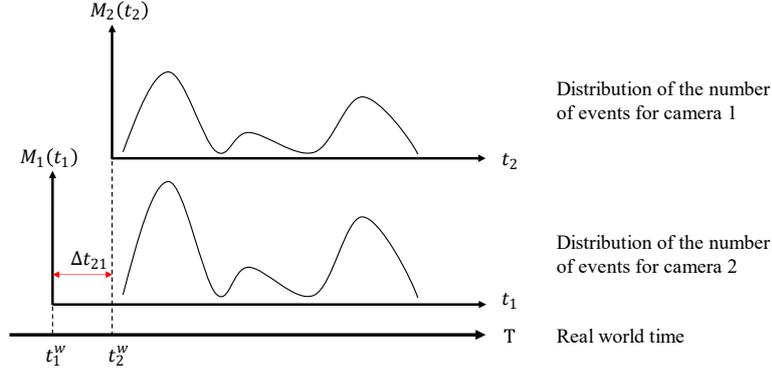

**Fig. 2.** An illustration of the different "time" in two event cameras and the real world. $t_1$ and $t_2$ are the running times for camera 1 and camera 2, and $T$ stands for world time. $t_1^w$ and $t_2^w$ are the start times for the two cameras. Owing to trigger and transmission delays, the event cameras do not start at the same moment, causing a misalignment in time between the cameras. The time offset is what we need to obtain to help adjust and synchronize camera timestamps to ensure temporal consistency across the camera streams. $M_1(t_1), M_2(t_2)$ are defined in Sec. 3.2.

To distinguish the "time" in different sensors and the real world, we denote the time by:

$$t_{time\ source}^{reference\ timeline}$$

which represents the time $t$ is from $time\ source$ and described in $reference\ timeline$. We assume two event cameras are used; thus, the $time\ source$ and $reference\ timeline$ can be 1, 2 and $w$ which represent the camera starting first, the camera starting later and the real world, respectively.

Specifically, as shown in Fig. 2, the start times for the two cameras are denoted as $t_1^w$ and $t_2^w$. The running times for camera 1 and camera 2 are represented as $t_1^1$ and $t_2^2$, respectively. The time interval between camera 1 and camera 2 can be defined as $\Delta t_{21} = t_2^w - t_1^w$.

We can determine the relationship between the running times of camera 1 and camera 2 using the equation $t_1^1 = t_2^2 + \Delta t_{21}$, i.e., $\Delta t_{21} = t_1^1 - t_2^2$. This is valid for all world times $t_w^w$, where $t_w^w = t_1^w + t_1^1 = t_2^w + t_2^2$. If the $camera\ index$ is the same as the $reference\ timeline$, the superscript is omitted for simplification. We emphasize that the proposed method can handle different numbers of event cameras, while in this section we only use two cameras for derivation.

## 3.2 Synchronization Algorithm

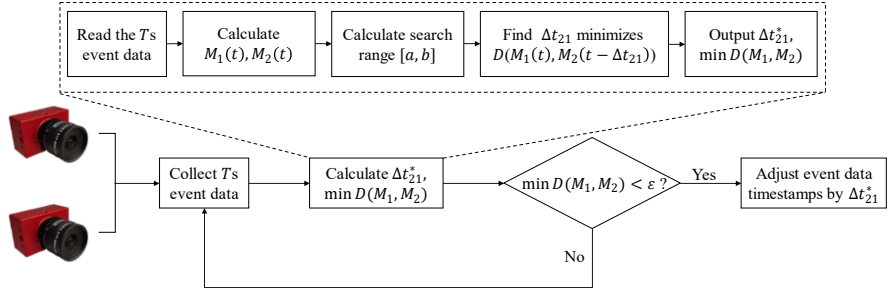

**Fig. 3.** Overall algorithm process. We calculate $\Delta t_{21}$ through $T$ s event data. If the $\min D(M_1, M_2)$ is less than the threshold $\varepsilon$, then we adjust the event data timestamps.

When the brightness change of a pixel in an image due to a moving edge (i.e., intensity gradient) exceeds the set threshold $C$, event data will be asynchronously outputted by the event camera. Each event contains four parameters, namely the pixel's X-coordinate $x$, Y-coordinate $y$, timestamp $t$, and polarity $p$.

The intensity of light at a point $(x, y)$ in the camera image at time $t$ is denoted as $L(x, y, t)$. Given a time interval $\Delta t$, the variance in the light intensity at that point can be represented as $\Delta L(x, y, t, \Delta t) = L(x, y, t + \Delta t) - L(x, y, t)$. As a result, the quantity of event data produced at that point within $\Delta t$ can be expressed as $n(x, y, t, \Delta t) = \left[\frac{\Delta L(x,y,t,\Delta t)}{C}\right]$, where [ ] denotes the largest integer less than or equal to that value. The number of event data generated by all pixels in $\Delta t$ is defined as $N(x, y, t, \Delta t) = \sum_{x=1}^{width} \sum_{y=1}^{height} n(x, y, t, \Delta t)$, where $width$ and $height$ refer to image dimensions. The total amount of event data generated over a longer period is represented as $\widetilde{N} = \sum_t N$. Assuming a uniform light source throughout the scene, and when $\Delta t$ is extremely small, any modifications in brightness within the image can be attributed solely to motion, allowing for an approximation [13]:

$$\Delta L \approx -\nabla \boldsymbol{L} \cdot \boldsymbol{v} \Delta t \tag{1}$$

which means the variance in light intensity $\Delta L$ can be attributed to a brightness gradient on the image plane denoted by $\nabla \boldsymbol{L}(x, y, t) = (\partial_x L, \partial_y L)$, moving at a velocity of $\boldsymbol{v}(x, y, t)$.

Assuming that the average gradient of the entire camera pixel image is represented by $\widetilde{\nabla \boldsymbol{L}}$ and that it satisfies the condition $\sum_{x=1}^{width} \sum_{y=1}^{height} \nabla \boldsymbol{L} = R \widetilde{\nabla \boldsymbol{L}}$. Similarly, the average velocity of the pixels $\widetilde{\boldsymbol{v}}$ should satisfy $\sum_{x=1}^{width} \sum_{y=1}^{height} \boldsymbol{v} = R\widetilde{\boldsymbol{v}}$, which leads to $\sum_{x=1}^{width} \sum_{y=1}^{height} \nabla \boldsymbol{L} \cdot \boldsymbol{v} \Delta t = R^2 \widetilde{\nabla \boldsymbol{L}} \cdot \widetilde{\boldsymbol{v}}$. $R$ denotes the total number of pixels present on the camera's image plane, i.e., $R = width \times height$. Consequently, the normalized event density indicating the quantity of events over a unit of time ($\tau = 1$ms) can be defined as follows:

$$M(t) = \frac{N(x,y,t,\tau)}{\widetilde{N}} = \frac{R^2}{\widetilde{N}} \left[ \frac{\widehat{\nabla L} \cdot \widetilde{v}}{C} \right], \ t = k\tau, k = 0,1,2 \ldots \quad (2)$$

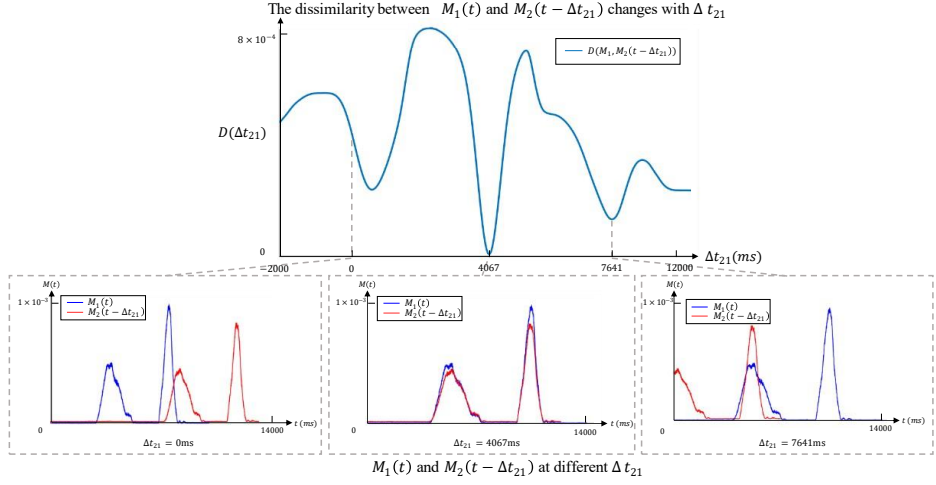

**Fig. 4.** An illustration of the $\Delta t_{21}$ traversal process. The upper figure represents the change of the dissimilarity between the two event density distributions over $\Delta t_{21}$, and the lower part is the event density distribution image corresponding to three $\Delta t_{21}$ we select. When $\Delta t_{21} = 4067$, the dissimilarity reaches the minimum value, which is the required $\Delta t_{21}^*$.

Two event cameras generally capture the same scene and have a similar average gradient. Thus, we assume $\widehat{\nabla L_1}(t_1) = \widehat{\nabla L_2}(t_2)$ for our deduction.( This assumption may not be satisfied all the time but the experiments in Sec.4.1 show that it does not have an obvious effect on synchronization results) If two cameras move together with a same speed (i.e., $\widetilde{v_1}(t_1) = \widetilde{v_2}(t_2)$) at any time, we have $M_1(t_1) = M_2(t_2)$ resulting in $M_1(t_w - t_1^w) = M_2(t_w - t_1^w - \Delta t_{21})$. Shifting a function left or right does not alter its shape, which leads to $M_1(t_w) = M_2(t_w - \Delta t_{21})$. In order to find the time difference between the start times of the two cameras, we need to find $\Delta t_{21}$, that minimizes the dissimilarity of the event density distributions of both cameras. That is, at any given time $t_w$, we seek $M_1(t_w) = M_2(t_w - \Delta t_{21})$. This problem is formulated as:

$$\Delta t_{21}^* = \arg \min_{\Delta t_{21}} D(M_1(t_w), M_2(t_w - \Delta t_{21})) \quad (3)$$

The function $D(M_1, M_2)$ quantifies the dissimilarity between $M_1$ and $M_2$ by utilizing the mean squared error, which is:

$$D(M_1(t_w), M_2(t_w - \Delta t_{21})) = \sum_{t_w} (M_1(t_w) - M_2(t_w - \Delta t_{21}))^2 \quad (4)$$

The initial $T$s of event data is used to determine the start time difference $\Delta t_{21}$ of the cameras. If $\min D(M_1, M_2)$ exceeds a threshold $\varepsilon$ during this period, the subsequent $T$s of event data will be used for further calculation until $\min D(M_1, M_2)$ less than $\varepsilon$ is obtained. $T$ is set to be 10 and $\varepsilon$ is set to be 0.0001 in our method. The timestamps of all outputs are adjusted by subtracting $\Delta t_{21}$ to compensate for the varying start times of the event cameras. The complete algorithm process is illustrated in Fig. 3.

As shown in Fig. 4, we traverse the $\Delta t_{21}$ and calculate the square of the difference of two functions with varying $\Delta t_{21}$ to minimize the dissimilarity. We set the search range to $[a, b]$. To determine $a$ and $b$, we calculate the difference between the corresponding percentiles of the two functions. In particular, we determine the timestamps $Q_1$ and $Q_2$ by calculating the $p^{th}$ percentiles of $M_1(t)$ and $M_2(t)$, respectively. Then we set $a = Q_2^p - Q_1^p - 2|Q_1^p - Q_2^p|, b = Q_2^p - Q_1^p + 2|Q_1^p - Q_2^p|$. Binary percentiles are used in this paper. Next, the range should be searched for the minimum value of $D(M_1, M_2)$ to determine the corresponding value of $\Delta t_{21}$.

## 4 Experiment

This Section begins by verifying the theoretical validity and accuracy of the proposed algorithm through experiments conducted under conditions that satisfy the algorithm's assumptions. Then, we test the algorithm in indoor and outdoor environments with two models of event cameras. Finally, we conduct experiments using two cameras that cannot be hardware-synchronized as well as multiple cameras of different models to demonstrate the actual performance of the algorithm.

### 4.1 Analysis of synchronization results with external triggers

Certain models of event cameras can receive electrical pulse signals via a hardware synchronization interface. These pulse signals can then be logged by the event cameras with their respective timestamps. Thus, the differences in the timestamps of identical signals received by several event cameras indicate the differences between their start times and serve as a reference value in our algorithm for evaluation. The synchronization error is the absolute value of the difference between $\Delta t_{21}^*$ and the reference value.

We perform experiments in a variety of scenarios to validate the algorithm. Since the algorithm relies on the uniformity of scene information captured by the event cameras, i.e., the same average image gradient, we select three scenarios that satisfy this assumption. Two DAVIS 346 cameras are used to conduct the experiments. The three scenarios are as follows: 1) recording human motion with stationary cameras, 2) capturing a pre-existing dataset displayed on a computer screen with stationary cameras, and 3) moving cameras to record a calibration chessboard. In scenario 1, a person performs a hand-raising motion at a distance of 2m from the cameras. For scenario 2, we select the dynamic_6dof sequence (dynamic six-degree-of-freedom motion) from Event Camera Dataset [14]. We export the data at a rate of 120 images per second due to the low frame rate of the original data and capture it using the event cameras. In scenario 3, we connect the cameras rigidly and move them at a speed of 10cm/s to record the calibration chessboard in front of a white wall. In each scenario, the object is captured within the consolidated field of view of the cameras, both of which are placed as close together as possible and facing the same direction to ensure consistency in the recorded scenes. We repeat each experiment 10 times, and each data sequence lasts 30s. We calculate the average, maximum, and standard deviation of synchronization errors for each experiment. The results are presented in Table 1.

**Table 1.** Synchronization errors in scenarios that satisfy the algorithm assumptions (units: ms)

| Scenes | Average | Maximum | Standard Deviation |
|---|---|---|---|
| Waving people (Fig.5(a)) | 2.27 | 2.34 | 0.06 |
| Computer screen (Fig.5(b)) | 1.69 | 2.95 | 0.72 |
| Chessboard (Fig.5(c)) | 1.39 | 1.96 | 0.42 |

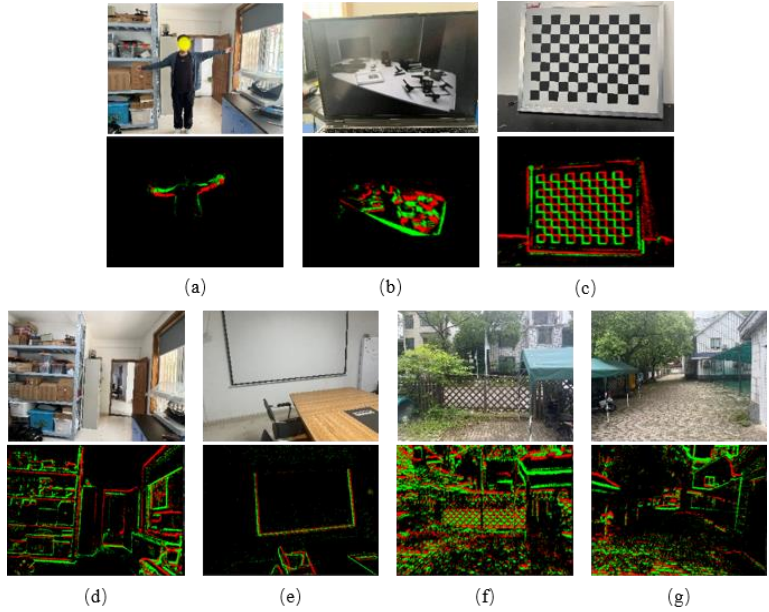

**Fig. 5.** Images of experimental scenes and corresponding event cameras' output. (a) people waving; (b) video on a computer screen; (c) chessboard in front of a white wall; (d) room with more objects; (e) room with fewer objects; (f) outdoor scene with near objects; (g) outdoor scene with far objects. For image (a-c), the gradient captured by the two cameras is the same, while image (d-g) represents a more general scene.

Experiments in a variety of indoor and outdoor environments are conducted to test the performance of the algorithm with both camera and human motion. Furthermore, we evaluate the algorithm's accuracy of different models of cameras, employing a DAVIS 346 camera and a Prophesee EVK1 Gen4 camera. In addition, we investigate the impact of camera parallel distances (20cm and 60cm) in different indoor and outdoor scenarios, with the camera either moving (at a speed of 10cm/s) or capturing human motion (in which the person raises one or both hands from a distance of 2m). These experiments aim to confirm the algorithm's effectiveness across common indoor and outdoor settings. It is worth noting that the assumption in Section 3 of equal average image gradient does not hold in these experiments. Table 2 presents the experimental settings and the corresponding results.

**Table 2.** Synchronization errors of a DAVIS 346 and a Prophesee EVK1 Gen4 (units: ms)

| Seq | Scenes | Descriptions | Average | Maximum |
|---|---|---|---|---|
| 1 | | Moving camera, 20cm baseline | 2.76 | 6.56 |
| 2 | Indoor 1 | Moving camera, 60cm baseline | 2.78 | 6.32 |
| 3 | (Fig.5(d)) | Dynamic people, 20cm baseline | 3.06 | 7.32 |
| 4 | | Dynamic people, 60cm baseline | 3.47 | 6.54 |
| 5 | | Moving camera, 20cm baseline | 2.52 | 4.75 |
| 6 | Indoor 2 | Moving camera, 60cm baseline | 3.43 | 7.51 |
| 7 | (Fig.5(e)) | Dynamic people, 20cm baseline | 3.15 | 6.58 |
| 8 | | Dynamic people, 60cm baseline | 3.63 | 6.88 |
| 9 | | Moving camera, 20cm baseline | 1.79 | 5.15 |
| 10 | Outdoor 1 | Moving camera, 60cm baseline | 1.88 | 5.63 |
| 11 | (Fig.5(f)) | Dynamic people, 20cm baseline | 2.03 | 5.36 |
| 12 | | Dynamic people, 60cm baseline | 3.88 | 7.91 |
| 13 | | Moving camera, 20cm baseline | 2.47 | 4.40 |
| 14 | Outdoor 2 | Moving camera, 60cm baseline | 3.14 | 6.92 |
| 15 | (Fig.5(g)) | Dynamic people, 20cm baseline | 2.37 | 4.14 |
| 16 | | Dynamic people, 60cm baseline | 3.23 | 4.69 |

We experiment in extreme scenarios that involve placing two cameras in reverse directions and low-light conditions. When the cameras are in reverse directions, the scenes recorded by each camera are entirely different. However, despite the different captured scenes, our algorithm can still align the event data because it requires the event density distributions instead of solely the captured images. Due to the camera's motion, events can still be obtained, resulting in similar event density distributions. Under low light conditions, the average illumination intensity is at 30 lux, while for the above indoor and outdoor environments, the average illumination intensity is 300 lux, and 10000 lux respectively. Since the event camera detects changes in illumination rather than relying on absolute illumination values, the event camera can still obtain events effectively. This is demonstrated by the results in Table 3.

**Table 3.** Synchronization errors in extreme environments (units: ms)

| Sensors | Conditions | Average | Maximum |
|---|---|---|---|
| DAVIS-DAVIS | Cameras Reverse placement | 3.46 | 7.72 |
| | Low light intensity | 3.27 | 5.35 |
| DAVIS-Prophesee | Cameras reverse placement | 4.29 | 9.42 |
| | Low light intensity | 4.57 | 8.25 |

The results indicate that the algorithm has small overall synchronization errors as the maximum average error under non-extreme conditions is 6.92ms. Cameras placed at a certain distance under the same motion condition have greater synchronization errors compared to closely placed cameras. Generally, the algorithm performs better in outdoor environments compared to indoor environments. The results obtained from the reverse experiment and the dim lighting experiment have larger synchronization errors, however, they are still within acceptable limits. Hence, the algorithm is effective under all these scenes.

### 4.2  Analysis of synchronization results without external triggers

The experiments are carried out involving CeleX5 cameras. The absence of hardware synchronization support in the cameras makes it impossible to verify synchronization outcomes and accuracy through mathematical means. A qualitative approach is employed instead to show the accuracy. The experimental setup involves an animated digital timer exhibiting minutes, seconds, and milliseconds separately displayed on a computer screen. Using Fig. 6 as an illustration, the computer screen is captured collectively by the event cameras. The computer screen refreshes at a rate of 120Hz which means that the screen changes approximately every 8.4ms.

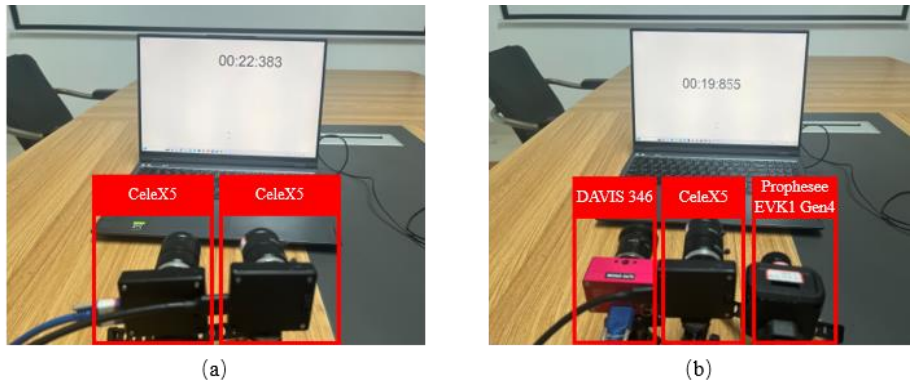

**Fig. 6.** An illustration of the experimental device. Event cameras capture the computer screen at the same time. (a) two CeleX5 cameras. (b) DAVIS 346, CeleX5, and Prophesee EVK1 Gen4 camera.

As shown in Fig. 6(a), the experiment uses software to simultaneously activate the two CeleX5 cameras and the timer. The output event data timestamps of the two cameras are plotted at the cameras' running time of 3s. The event frame [5] drawing period is set at 8ms. The results are shown in Fig. 7. The first camera displays the time of 00:06:753, indicating that the camera actually starts when the timer is at 3.753s due to the delay. The time displayed by the second camera is 00:04:348, stating that the second camera starts to collect data 1.348s after the timer starts. The synchronization of the two cameras using the algorithm produces outputted images that both show a time of 00:06:753. We state that it does not mean that the algorithm achieves 1ms accuracy here, but the error is less than the timer refreshing cycle, i.e., 8.4ms.

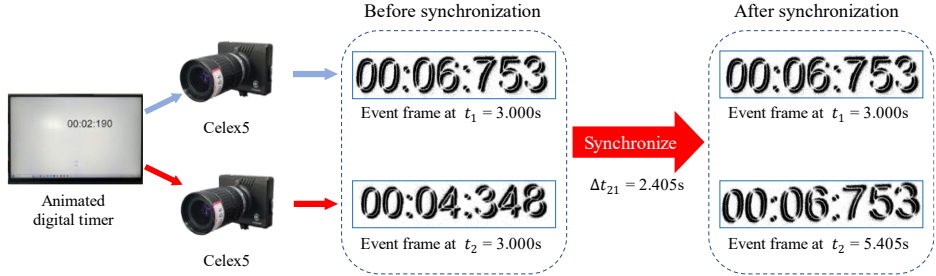

**Fig. 7.** Experimental results of two CeleX5 cameras. We draw the event frames of two event cameras with a running time of 3.000s, and the digital difference displayed in the image can be used to represent the difference between the start time. After synchronization, the displayed time was the same, proving that the synchronization error was less than 8.4ms.

Then, we conduct a similar experiment using a DAVIS 346 camera, a CeleX5 camera, and a Prophesee EVK1 Gen4 camera to validate the synchronization of multiple cameras of different models, as shown in Fig. 6(b). The results are shown in Fig. 8, which also proves that the synchronization error of the algorithm is within 8.4ms.

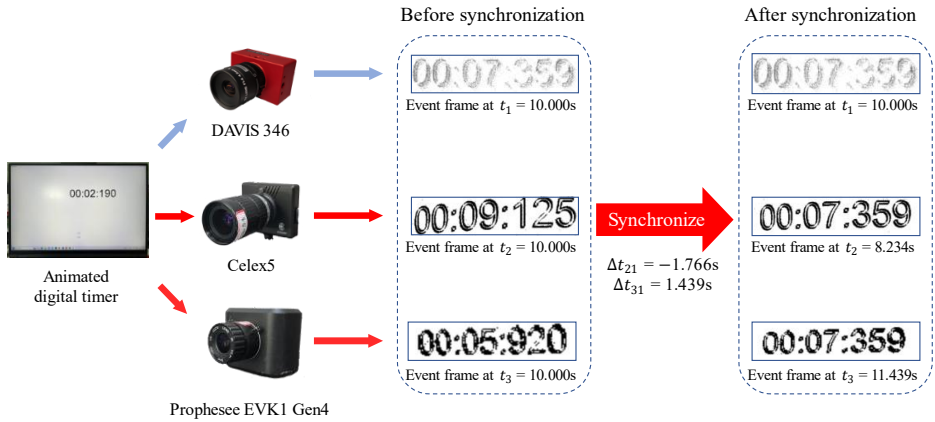

**Fig. 8.** Experimental results of a DAVIS 346 camera, a CeleX5 camera, and a Prophesee EVK1 Gen4 camera.

## 5 Conclusion

In this paper, we present a hardware-free time synchronizing method to synchronize different event cameras in a multi-camera system. The proposed method calculates the event density distributions of each event camera and estimates their start delays by minimizing the dissimilarity between the distributions. Experiments show that synchronization errors are within 10ms in indoor, outdoor, and even challenging scenes. The method can handle the problem of synchronizing any event cameras regardless of models and numbers.